\documentclass{bmvc2k}

\usepackage{adjustbox}
\usepackage{enumitem}
\usepackage{bm}

\usepackage[nodisplayskipstretch]{setspace}

\usepackage{siunitx}

\usepackage{amsfonts}

\usepackage{graphicx}

\usepackage{multirow}

\usepackage{array}

\usepackage{makecell}

\usepackage{verbatim}

\usepackage{amsmath}
\usepackage{tabularx}

\newcommand{\etal}{\textit{et al}. }
\newcolumntype{?}[1]{!{\vrule width #1}}
\newcolumntype{P}[1]{>{\centering\arraybackslash}p{#1}}

\title{TARN: Temporal Attentive Relation Network for Few-Shot and Zero-Shot Action Recognition}

\addauthor{Mina Bishay}{m.a.t.bishay@qmul.ac.uk}{0}
\addauthor{Georgios Zoumpourlis}{g.zoumpourlis@qmul.ac.uk}{0}
\addauthor{Ioannis Patras}{i.patras@qmul.ac.uk}{0}

\addinstitution{
Multimedia and Vision Research Group \\
Queen Mary University of London \\
London, UK
}

\runninghead{BISHAY, ZOUMPOURLIS, PATRAS}{Temporal Attentive Relation Network}

\def\etal{\emph{et al}\bmvaOneDot}

\begin{document}
\maketitle
\begin{abstract}
In this paper we propose a novel Temporal Attentive Relation Network (TARN) for the problems of few-shot and zero-shot action recognition. At the heart of our network is a meta-learning approach that learns to compare representations of variable temporal length, that is, either two videos of different length (in the case of few-shot action recognition) or a video and a semantic representation such as word vector (in the case of zero-shot action recognition). By contrast to other works in few-shot and zero-shot action recognition, we a) utilise attention mechanisms so as to perform temporal alignment, and b) learn a deep-distance measure on the aligned representations at video segment level. We adopt an episode-based training scheme and train our network in an end-to-end manner. The proposed method does not require any fine-tuning in the target domain or maintaining additional representations as is the case of memory networks. Experimental results show that the proposed architecture outperforms the state of the art in few-shot action recognition, and achieves competitive results in zero-shot action recognition.
\end{abstract}


\section{Introduction}
\label{sec:intro}

Human action recognition has received significant attention in the last decade due to its application in areas like video surveillance, human-computer interaction, and video retrieval \cite{herath2017going}, as is the trend in most Computer Vision problems, using Deep Neural Networks \cite{tran_15_c3d, donahue_17_lrcn, simonyan_14_twostream, palasek2017discriminative}. However, training deep architectures requires a large amount of annotated data, something that is not easily available for new action classes. By contrast, humans are able to recognize new actions using a few labelled examples or just an action-related description.

For this reason, Few-Shot Learning (FSL) and Zero-Shot Learning (ZSL) have recently received a lot of attention. Most of the FSL works follow the meta-learning approach where a high-level transferable knowledge is learned on a collection of different tasks. This knowledge, that helps to perform classification on the target few-shot task(s), can be good initial network weights \cite{finn_17_maml}, embedding functions \cite{vinyals_16_oneshot, snell_17_prototypical}, or an external memory with useful information ~\cite{munkhdalai_17_meta,santoro_16_memory}. Some works treat the few-shot problem as a similarity problem, that is, a similarity model is trained to classify a query example by comparing it to labelled examples in the training set. These works are simple and need no additional memory or cost, however, most of them used a fixed distance metric to calculate the matching score. In \cite{sung_18_compare}, Sung \etal proposed a relation network that learns to calculate embeddings and a transferable deep measure of similarity between them. \cite{sung_18_compare} achieved state-of-the-art results in image-based FSL, and shows competitive performance when applied to image-based ZSL.

However, most of the works have focused on image-based FSL and ZSL problems, like object recognition \cite{finn_17_maml, snell_17_prototypical, sung_18_compare, zhang2017learning}, while relatively few works were aimed at video-based FSL and ZSL problems like action recognition \cite{xu2016multi, wang_17_bidilel, zhu_18_compound}. Applying FSL and ZSL in videos is more challenging compared to images, due to the additional temporal dimension in videos and the variations that are introduced.
To the best of our knowledge only one work has been proposed for few-shot action recognition~\cite{zhu_18_compound}. However, this work is based on memory networks that require extra computational and space resources, and use a single embedding vector to represent the entire video. This might not capture well the temporal structure of the action and is challenging due to the amount of information existing in it. Xu \etal~\cite{xu2018dense} proposed a method for action recognition from videos in limited data scenarios (i.e. when only a portion of the examples is available), however, training and testing is done on all the classes which is different from the few-shot protocol of~\cite{vinyals_16_oneshot} that is typically used in such problems. All ZSL approaches for action recognition use a single vector as a visual representation of entire videos, obtained either from handcrafted~\cite{zhu_2018_cvpr} or from deep~\cite{mishra_18_generative} features. Working on video level, thus being incapable to explicitly leverage time-specific information, they miss more detailed visual cues that appear on the fine-grained level of the segments that form a video.

Our network (TARN) addresses the few-shot problem by working at video-segment level to calculate the relation scores between a query video and other sample videos -- the query video is then assigned with the label of the most related video in the sample set. The relation/similarity is calculated in two stages: the embedding stage and the relation stage. In the embedding stage, a C3D~\cite{tran_15_c3d} network followed by a layer of bidirectional Gated Recurrent Unit (GRU)~\cite{cho_14_gru}, extract features from short segments of videos. The GRU learns an embedding function that is general and transferable over different tasks. In the relation stage, a segment-by-segment attention mechanism is used to align segment embeddings for a pair of query and sample videos, and then the aligned segments are compared. That is, we introduce segment-to-segment comparisons and model their temporal evolution, as a prior step towards video-to-video matching. The comparison outputs are fed to a deep neural network that learns a general deep distance measure for video matching, and gives at its output the final relation score for a pair of videos. Our approach can generalise to video-based ZSL, where no videos are available for the training in the sample set, but instead class descriptions (e.g. attribute vectors) are given. In this case, the query and the sample set have different types of data (visual and semantic, respectively) and therefore we use two different embedding modules, one for each domain. TARN (excluding the C3D model) is trained in an end-to-end manner in both FSL and ZSL cases. The main contributions of our work are three-fold:
\begin{enumerate}
[noitemsep]
\item We propose a relation network for few-shot and zero-shot action recognition. The proposed architecture compares either segment-wise visual features from a pair of videos (in FSL), or segment-wise visual features from a video with a class-wise semantic representation (in ZSL), retaining temporal information to finally perform video-wise classification.
\item The proposed architecture needs no additional resources like memory networks and does not require training or fine-tuning on the target problem like \cite{finn_17_maml, ravi_17_optimization}
\item We test the proposed architecture on different benchmark datasets, achieving state-of-the-art results in FSL and very competitive performance in ZSL.
\end{enumerate}
\section{Related work}

Our approach aims to tackle the problems of few-shot and zero-shot action recognition, that are related to the following research directions:

\textbf{Few-shot learning}:
Since the seminal work of Fei-Fei \etal in \cite{feifei_06_oneshot}, several works have been proposed on learning from few examples, focusing mostly on image classification. In \cite{finn_17_maml}, Finn \etal addressed the problem of FSL by focusing on fast adaptability through proper initialization conditions. The danger of overfitting on few-shot tasks when adopting finetuning~\cite{ravi_17_optimization} has been noted in \cite{vinyals_16_oneshot}. A remedy to this has been the episode-based strategy of \cite{vinyals_16_oneshot}, where at each episode \textit{$K$} support examples from each one of \textit{$C$} classes (where $K$ and $C$ are typically small), and one query example, are randomly chosen and the network weights are updated according to a loss defined over them.
In this way, the generalization is improved without the need to perform weight updates on the support set during inference. Other works~\cite{koch2015siamese, snell_17_prototypical, vinyals_16_oneshot} treated FSL as a metric-learning problem, where the query samples are classified using either pre-defined or learned distance measures on learned embeddings. 
The closest work to our FSL method is ~\cite{sung_18_compare}, where query images are classified by comparing them to images from the support set. In \cite{sung_18_compare}, a relation module learns a non-linear similarity function to match images. In our work, we propose a relation module that first uses Euclidean distance and cosine similarity for comparing video segments, and then a trainable deep network for modeling the temporal distances/similarities across different segments and inferring a relation score for each pair of videos, is subsequently learned.

\textbf{Zero-shot learning} was initially defined as a problem in the works of Palatucci \etal\cite{palatucci_09_zeroshot} and Larochelle \etal\cite{larochelle_08_zerodata}. Action recognition in the ZSL scenario typically requires bridging the semantic gap between the distributions of the semantic representations and the visual representations from the unseen classes~\cite{wang_17_bidilel, kodirov2015unsupervised, xu2015semantic}. The semantic representation of a class is a single vector, which can be either class-related attributes or word2vec embedding of the class label~\cite{liu_11_attributes,mishra_18_generative}. The visual representations used in existing ZSL approaches are either handcrafted~\cite{zhu_2018_cvpr,qin2017zero,xu2016multi} based on the Improved Dense Trajectories (IDT) method~\cite{wang2013action}, or deep~\cite{mishra_18_generative,wang_17_bidilel,GanYG16} features extracted with C3D~\cite{tran_15_c3d} network. IDT features represent a video with a single vector, by default. C3D features refer to 16-frame video segments, but most of the ZSL methods average them over the entire video to obtain a single visual video representation. In ZSL, the embeddings of the semantic representations are mapped to the embeddings of the visual representations, to establish relationships between the class-related attributes and the visual features. In \cite{gan_15_sir}, a ZSL method was proposed, leveraging the inter-class semantic relationships between the known and unknown actions. Other works have investigated using auxiliary data that are relevant to the unseen classes \cite{xu2016multi}, using Wikipedia as an external ontology, to calculate semantic correlations between class labels \cite{gan_16_recognizing}, or constructing universal representations to achieve cross-dataset generalization \cite{zhu_2018_cvpr}. However, even in datasets of trimmed videos, factors such as camera motion, viewpoint or action structure complexity can obstruct the extraction of meaningful features at some video parts. Hence, building deep networks that are capable of processing segment-level visual features, is a promising direction for ZSL.

\textbf{Sequence matching}:
We claim that matching video sequences can benefit from comparing video segments as a first step. Previously, Fernando \etal\cite{fernando_17_unsupervised} have proposed a method to match video segments from a pair of videos, that share similar temporal evolutions. We also draw our inspiration from works on text matching, where representations of sequence parts are compared and aggregated to match an entire text sequence~\cite{wang_17_match}. An attention mechanism similar to that of ~\cite{bahdanau_15_attention, rocktaschel_16_attention} is used to align the segments of each video in the support set, with respect to the query video. This is done to semantically align the features of the support set videos to the features of the query video. It also transforms the number of segments for each video of the support set to be equal to that of the query video. 

\section{Proposed Architecture}

\begin{figure}
  \includegraphics[width=1\linewidth]{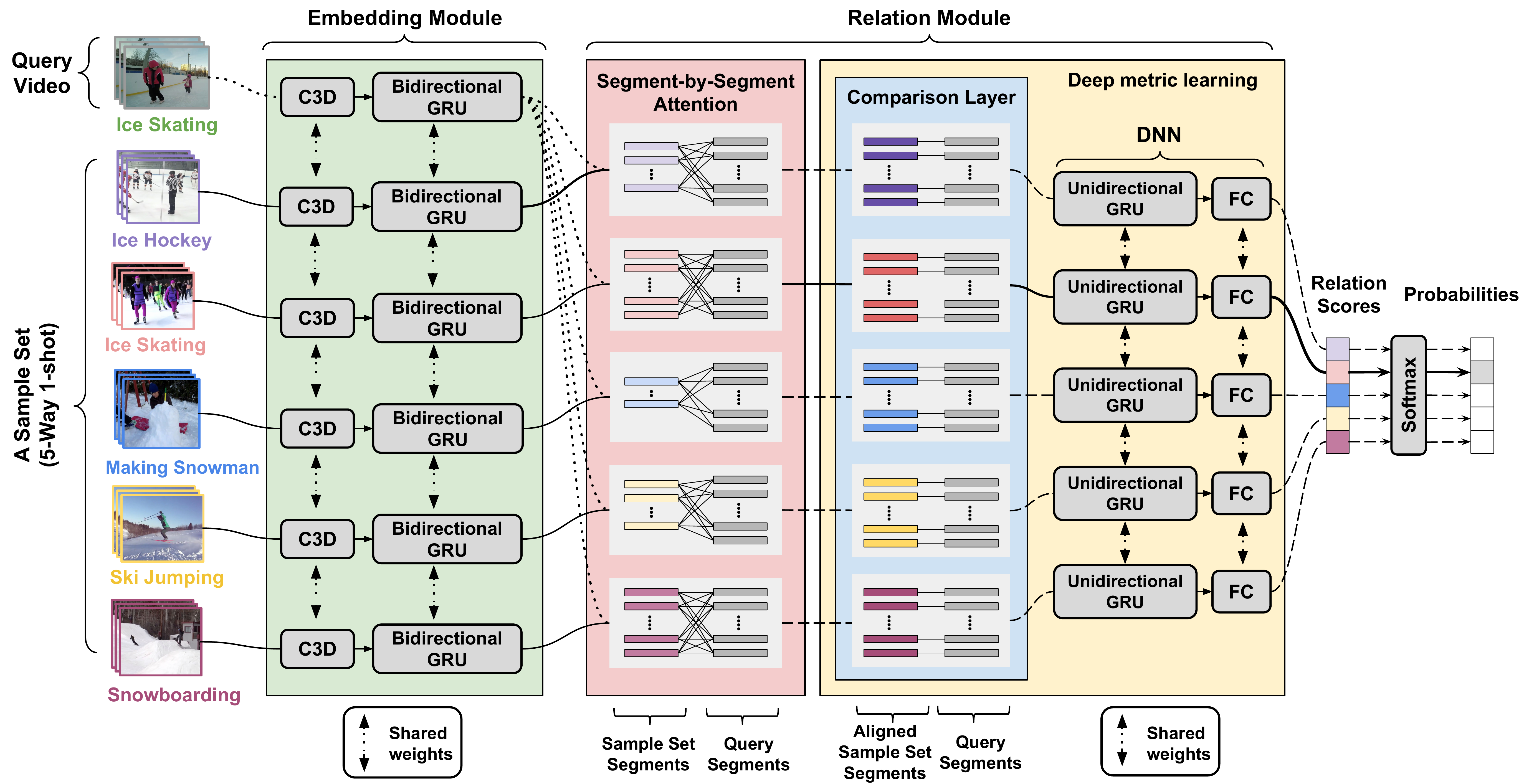}
  \caption{The proposed TARN architecture, consisting of the embedding module and the relation module. In an \textit{C}-way \textit{K}-shot task (where $K>1$), the relation score of the query video to each class of the support set, is the average of the sample relation scores of that class.}
  \label{fig:tarn_model}
\end{figure}

In this section we introduce a novel deep architecture, named Temporal Attentive Relation Network (TARN) for the problems of Few-Shot Learning (FSL) and Zero-Shot Learning (ZSL) for video-based tasks. Figure~\ref{fig:tarn_model} shows an overview of the network. TARN learns to compare a query video against a sample set of videos in FSL, or semantic attributes in ZSL, representing a group of actions. In the FSL case, the inputs are segment-wise visual representations of the query and sample videos, and in the ZSL case the inputs are segment-wise visual representations of a query video on the one hand and semantic representations of the unseen classes on the other. The output is a relation score, either for each pair of videos, or for each pair of video and semantic attributes. More specifically, TARN consists of two modules: the embedding module and the relation module. First, the embedding module processes the visual or semantic representations, retaining the temporal structure of visual features, and produces embeddings that are later compared. Second, the relation module initially applies segment-by-segment attention. By doing so, it either transforms the visual representations of the sample set to have the same number of segments as the query video (FSL), or transforms the semantic representations of the unseen classes to allow a segment-wise comparison between them and the visual representations (ZSL). Afterwards, a per-segment comparison is performed. Finally the relation module produces the matching score by taking into account the variations of the comparisons across all segments of the query video. These modules are explained in detail in the following subsections.

\subsection{Embedding Module}

TARN uses a single embedding module to embed both the query and sample videos in FSL, while two different embedding modules are used in ZSL. That is, one is used to embed the visual data while the other is used to embed the semantic data. 

\noindent \textbf{Video embedding:} A pre-trained C3D network is used to extract spatio-temporal features across short segments of videos. This network acts as a local feature extractor in the embedding module. Then, a bi-directional GRU uses the local features to learn more globally-aware features, allowing each time step to access both backward and forward information across the whole video. Moreover, it reduces the dimension of the C3D features. This visual embedding module is applied in the same way to embed video segments in both the FSL and ZSL cases.

\noindent \textbf{Semantic embedding:} ZSL first compares visual representations extracted from segments of the query video to semantic representations of the sample set classes. The original semantic information needs to be encoded into a representation that allows a feature-rich comparison of video segments to class-related attributes, and therefore needs to have the same dimensions with the visual representation of the video segments. The semantic embedding module consists of two stacked fully-connected (FC) layers.
\subsection{Relation Module}

In this section we will explain the relation module from the FSL perspective, and then we will show how it is extended for ZSL. In FSL, in order to match a query video to a sample set of videos, firstly we pair the query video with each video in the sample set. Given pairs of videos, secondly we align segments in the videos using a segment-by-segment attention layer. The attention layer maps the sample video to have the same number of segment embeddings as the query. Third, each segment in the query is compared to the corresponding aligned sample segment. Fourth, the comparison outputs of the different segments are fed to a deep neural network, that learns a deep metric for video matching, and gives at its output a relation score for each pair. Finally, a softmax layer is used to map the relation scores to a probability distribution over the sample classes. 


\textbf{Segment-by-segment attention:} Several recent works in text sequence matching and textual entailment use an attention mechanism, named word-by-word attention, to align the words of two given sentences~\cite{bahdanau_15_attention, parikh_etal_2016_decomposable, rocktaschel_16_attention, wang_17_match}. Similarly, as shown in the corresponding block of Fig.~\ref{fig:tarn_model}, we adopt the word-by-word attention in our architecture to align the sample and query segment-embeddings (i.e. segment-by-segment attention). Given a sample video $\bm{S} \in \mathbb{R}^{N \times d}$ and a query video $\bm{Q} \in \mathbb{R}^{M \times d}$, where each row in $\bm{S}$ and $\bm{Q}$ represents a segment-embedding vector of dimension $d$, and where $N$ and $M$ denote the number of segments in videos $\bm{S}$ and $\bm{Q}$ respectively. The segment-by-segment attention is calculated as follows:
\begin{equation}
\label{eq:weight_matrix}
\bm{A}=softmax((\bm{SW} + \bm{b} \otimes \bm{e}_{N}) \bm{Q}^T), \phantom{.......} \bm{H = A^T S},
\end{equation}
where $\bm{W} \in \mathbb{R}^{d\times d}$ and $\bm{b} \in \mathbb{R}^{d}$ are parameters to be learned, and the operator ``$\otimes \bm{e}_{N}$'' repeats the bias vector $\bm{b}$, $N$ times to form a matrix of dimension $N\times d$. $\bm{A}  \in \mathbb{R}^{N\times M}$ is the attention weight matrix and $\bm{H}$ is the aligned version of $\bm{S}$. Each row vector in $\bm{H}$ is a weighted sum of the $\bm{S}$ segment-embeddings, and represents the parts of $\bm{S}$ that are most similar to the corresponding row vector (segment-embedding) of $\bm{Q}$. The row vectors of $\bm{Q}$ and $\bm{H}$ are used as inputs to a comparison layer.


\textbf{Deep metric learning:} The relation module performs deep similarity/distance metric learning by using a comparison layer and a non-linear classifier on the top of it. The comparison layer calculates a similarity measure between each of the $M$ segments of the row vectors of $\bm{Q} \in \mathbb{R}^{M \times d}$ and $\bm{H} \in \mathbb{R}^{M \times d}$. This measure, as described in ~\cite{wang_17_match}, can be based on one of the following operations: multiplication (Mult), subtraction (Subt), neural network (NN), subtraction and multiplication followed by a neural network (SubMultNN), or Euclidean distance and cosine similarity (EucCos). Since the measure is estimated at each of the $M$ pairs of segments, the output of this layer has also $M$ dimensions.
This layer acts as an intermediate stage that produces low-level representations of the comparisons between the sample and the query segments. As we will show in the experimental section, decomposing the query-sample matching problem into several comparisons across segments performs better than just a single comparison of two vectors representing the sample and query videos~-- this coincides with the findings in text sequence matching problems~\cite{wang2016machine, he2016pairwise, parikh_etal_2016_decomposable}.

Unlike other works in FSL which used a linear classifier or a fixed metric to match query and sample examples \cite{koch2015siamese, snell_17_prototypical}, we follow \cite{sung_18_compare} and use a deep neural network for deep metric learning. That is, the outputs of the comparison layer are passed to the deep network that learns a global deep metric over the entire videos. For FSL, we use a uni-directional GRU to learn temporal information across different segment comparisons and a Fully-Connected (FC) layer for giving the final relation score. The final relation scores coming from different sample examples are passed to a softmax layer, so that they can be mapped to a probability distribution over the sample classes. The query video is assigned with the label of the most related video in the sample set. In the case of multiple shots ($K>1$) per class in the sample set, the mean of the relation scores over the shots of each class is taken as the relation score of the query video with that class.

The full architecture (excluding the C3D model) is trained in an end-to-end fashion. We use episode-based training scheme proposed in \cite{vinyals_16_oneshot, snell_17_prototypical} to train our architecture, and binary cross-entropy as the cost function. The total batch/episode cost is:
\begin{equation}
L(t,q) = - \dfrac{1}{KC} \sum_{k=1}^{K} \sum_{c=1}^{C} (t_{kc} \log q_{kc} + (1-t_{kc}) \log (1-q_{kc}))
\end{equation}
where $K$ denotes the number of shots, $C$ the number of classes in each episode, $t$ the target relation score, and $q$ the predicted relation value.

In \textbf{zero-shot learning}, each of the sample set classes is expressed by a semantic attribute/word vector.  In this case, the attention mechanism estimates the similarity between the semantic vector and each segment of the query video, instead of aligning video segments as in the FSL case. The semantic embedding module encodes class attributes in representations that allow finding relations between the underlying class attributes and query video segments. We perform several segment-to-attribute comparisons, leveraging this fine-grained information to improve the training process and reduce overfitting. The comparison outputs are aggregated over all query video segments using two FC layers and an average pooling layer that produces the final relation score. The comparison, FC, and pooling layers represent in this case the deep network for metric learning. Note that the uni-directional GRU of the FSL case has been replaced, as there is no temporal alignment information to learn in the deep network.
\section{Experimental results}
\label{Exp_results}
\subsection{Few-shot action recognition}


\textbf{Implementation details.}  In \cite{zhu_18_compound}, Zhu and Yang introduced a dataset for few-shot video classification, that is a modification of the original Kinetics dataset \cite{kay_17_kinetics}. In this work, we follow \cite{zhu_18_compound} and use their dataset and evaluation protocol. The dataset videos are sliced into fixed length segments of 16 frames each, and then these segments are fed to a C3D network pre-trained on Sports-1M \cite{karpathy_14_sports1m}. Visual features are extracted from the last FC layer (i.e. FC7) of the C3D network, and used as input to the embedding bidirectional GRU. The dimension of the C3D features is 4096, and the GRU has a hidden state of size 256. In the relation module, the output of the comparison layer is used as input to a unidirectional GRU layer of size 256, and the GRU output at the last time step is fed to an FC layer with a single neuron for predicting the final relation score. We use $20,000$ episodes for training, $500$ for validation, and $1,000$ for testing. Each episode has a sample set of 5 classes. We evaluate the performance of our architecture on the validation set every $500$ training episodes. The best-performing model on the validation set is used for testing. We train our architecture using Stochastic Gradient Descent (SGD) with momentum $m = 0.9$, and learning rate equal to $10^{-3}$ for the first $10k$ training episodes, and $10^{-4}$ for the last $10k$ episodes.

\textbf{Results.} In our first experiment, we investigate the impact of the various functions that can be used as distance/similarity measure in the comparison layer, on the TARN performance. Following \cite{wang_17_match}, we compare five different distance measures (Mult, Subt, NN, SubMultNN, EucCos). Table~\ref{FSL_Metrics} shows the accuracy obtained by the TARN model over the different measures. EucCos leads to the best accuracy over all shots. Although EucCos is a fixed measure with no learnable parameters, the following layers in the relation module are trainable and non-linear.


\begin{table}[!b]
\centering
\begin{adjustbox}{width=0.73 \textwidth,center}
\begin{tabular}{|P{1.0cm}|P{2.0cm}?{0.4mm}P{1.0cm}|P{1.0cm}|P{1.0cm}|P{1.0cm}|P{1.0cm}|}
\hline
Method     & Measure    & 1-shot & 2-shot & 3-shot & 4-shot & 5-shot  \\
\Xhline{3\arrayrulewidth}
\textbf{TARN}        & Mult\color{white}{*} & 63.10 & 71.16 & 74.08 & 76.36 & 75.64  \\
\hline
\textbf{TARN}        & Subt\color{white}{*} & 64.82 & 70.70 & 73.90 & 76.26 & 77.54  \\
\hline
\textbf{TARN}        & NN\color{white}{*} & 63.26 & 70.46 & 72.70 & 75.62 & 75.58  \\
\hline
\textbf{TARN}        & SubMultNN\color{white}{*} & 66.10 & 73.74 & 75.44 & 77.08 & 78.20  \\
\hline
\textbf{TARN}        & EucCos\color{white}{*} & \textbf{66.55}  & \textbf{74.56}  & \textbf{77.33}  & \textbf{78.89}  & \textbf{80.66}  \\
\hline
\end{tabular}
\end{adjustbox}
\caption{TARN model accuracy when using different similarity/distance measures in the comparison layer.
}
\label{FSL_Metrics}
\end{table}


In the next experiment, we investigate the benefits of using segment-by-segment attention and comparing segment-wise representations in our architecture. To do so, we compare the TARN model to another model that has no attention layer and performs a single comparison for each video pair. Specifically, we modify the embedding module by replacing the bidirectional GRU with a unidirectional GRU of size 256, and the relation module by replacing the unidirectional GRU with a FC layer of size 256 for deep metric learning. In this case, the embedding GRU output at the last time step summarizes the entire video into a single vector. We call this model ``TARN-single''. Table ~\ref{FSL_SOA} shows the obtained accuracies by both the TARN and TARN-single models. Aligning video segments through attention and comparing segment-wise embeddings leads to consistent gains over the different shots. Furthermore, TARN model achieves better accuracy than the state-of-the-art method \cite{zhu_18_compound}. The accuracy gains hold for all shots, with significant boosts in the more difficult one-shot case, showing that even with a single sample per class in the sample set, the TARN model can still perform well.




In the last experiment, we compare the TARN model to the architecture proposed in \cite{zhu_18_compound} when using the same visual features. In the comparison, we use features extracted from a ResNet-50 \cite{he_16_resnet} model pretrained on ImageNet \cite{russakovsky_15_imagenet}. ResNet features are extracted every 3 frames. We modify the TARN model to use the frame-level ResNet-50 features instead of the C3D ones, by replacing the bidirectional GRU in the embedding module by a unidirectional GRU, that is applied over video segments and gives an output at the last time step of each segment. The rest of the network remains the same to that of the original TARN model. This model is called  ``TARN-f''. In this comparison, we also show the effect of not using attention and segment-level comparisons. To do so, the unidirectional GRU is applied over the entire video, and then the GRU output at the last time step is fed to the comparison layer. This model is called ``TARN-f-single''. Table ~\ref{FSL_SOA} shows the results obtained by both the TARN-f and TARN-f-single models. First, we can see that TARN-f performs better than CMN \cite{zhu_18_compound} when using the same visual features (ResNet-50). Second, using attention and segment-wise comparisons improves the performance of our architecture. Finally, the C3D features perform better than the ResNet features, when trying to compare video segments in a few-shot scenario. This is probably due to the fact that C3D features are spatiotemporal, while ResNet features are static.

\begin{table}[!t]
\centering
\begin{adjustbox}{width=0.86 \textwidth,center}
\begin{tabular}{|l|P{2.0cm}?{0.4mm}P{1.0cm}|P{1.0cm}|P{1.0cm}|P{1.0cm}|P{1.0cm}|}
\hline

Method     & Features    & 1-shot & 2-shot & 3-shot & 4-shot & 5-shot  \\
\Xhline{3\arrayrulewidth}

\textbf{CMN~\cite{zhu_18_compound} - ECCV 2018} & \multirow{3}{*}{ResNet-50} & 60.5   & 70.0   & 75.6   & 77.3   & 78.9   \\ \cline{1-1} \cline{3-7} 

\textbf{TARN-f-single} &  &  57.92  &  63.96  &  65.90  &  68.52  &  70.64  \\ \cline{1-1} \cline{3-7} 

\textbf{TARN-f} &  & 64.83  & 72.94  & 76.22  & 78.02  & 78.52  \\ \hline

\textbf{TARN-single} & \multirow{2}{*}{C3D} & 62.84 & 69.80 & 73.96 & 74.88 & 76.88 \\  \cline{1-1} \cline{3-7} 

\textbf{TARN}        &   & \textbf{66.55}  & \textbf{74.56}  & \textbf{77.33}  & \textbf{78.89}  & \textbf{80.66}  \\  \hline

\end{tabular}
\end{adjustbox}
\caption{Accuracies of the the state-of-the-art method \cite{zhu_18_compound}, as well as the TARN model at different settings.}
\label{FSL_SOA}
\end{table}

\subsection{Zero-shot action recognition}
\textbf{Datasets and settings:} We use two action recognition datasets to evaluate our architecture, namely UCF-101 \cite{soomro_12_ucf101} and HMDB51 \cite{kuehne_11_hmdb51}. UCF-101 has 13320 video clips from 101 classes, while HMDB51 has 6766 clips from 51 classes. Following \cite{wang_17_bidilel}, we divide the 101 actions in UCF-101 into 51/50 and 81/20 (seen/unseen) classes. For HMDB51, we divide the 51 actions into 26/25 classes. In the literature, different numbers of splits (ranging between 5-50) are randomly generated for the seen/unseen classes, and the mean accuracy and standard deviation are reported over them. In this work, we follow \cite{wang_17_bidilel} and use 30 random splits used by \cite{xu2015semantic} for the 51/50 and 81/20 cases in UCF-101, and for the 26/25 case in HMDB51.


\textbf{Video/class representations:} In our experiments, we use two types of semantic representations, both of which are widely used in the literature. First, we use the 115 binary semantic attributes (denoted as ``Attr'') that are manually annotated by \cite{THUMOS14} for UCF-101. To the best of our knowledge, there are no semantic attributes available for HMDB51. Second, we use 300-dimensional Word Vectors (mentioned as ``WV'') generated by the skip-gram model of \cite{mikolov2013distributed}, that is trained on the Google News dataset. While there are other ways of representing semantic information, an extensive analysis of their influence is beyond the scope of this work. For example, \cite{qin2017zero} showed better performance using Error-Correcting Output Codes (ECOC). Here, we follow the majority of the works and use attributes and/or word vectors. Similar to FSL, we use C3D for extracting visual features.


\textbf{Implementation details:} The C3D features are embedded using a bidirectional GRU layer with a hidden state of size $256$. Hence, the output dimension of the GRU at each time step (i.e. video segment) is of size $512$. The semantic information is embedded using two FC layers with $4096$ and $512$ nodes. The deep network used for metric learning has two FC layers of size $256$ and $1$. We train our architecture in an end-to-end fashion using episode-based training strategy \cite{vinyals_16_oneshot, snell_17_prototypical}. Adam optimizer \cite{kingma2014adam} with an initial learning rate set to $10^{-4}$ and gradient clipping to $0.5$ is used in the training. The architecture is trained for $3,000$ episodes and tested for $100$ episodes. During training, episodes/batches of size $16$ for UCF-101 and $8$ for HMDB51 are used, while in testing episodes are formed from all the unseen classes in the target split. We evaluate our architecture every $50$ training episodes.


\textbf{Ablation studies:} In Table~\ref{TARN_ZSL} we show the results obtained by our architecture on ZSL using different settings. The first and second settings (first two rows in Table~\ref{TARN_ZSL}) show the performance of our architecture when having a single representation for an entire video, instead of multiple segment representations. In the first setting, we use a unidirectional GRU to summarize the video segments into a single vector, obtained from the last time step of the GRU, and perform a single comparison between the semantic and the video vectors. In the second setting, we use a bidirectional GRU that gives an output at each time step. The attention mechanism is applied to the visual and semantic embedding, so as to map the multiple segment representations of the query video into a single representation. The single aligned query representation is then compared to the semantic embedding. The third setting (third row in Table~\ref{TARN_ZSL}) is a network that performs per-segment comparisons between the visual and semantic features. The attention mechanism maps the single semantic embedding of each class into multiple representations, as many, as the number of segments in the query video. In this way, each segment of the query video is compared to a semantic representation, allowing us to find relations between class-related semantic attributes and the visual features of each segment. Regarding the single comparison cases, the results show that the attention mechanism (second row in Table~\ref{TARN_ZSL}) effectively encodes multiple segment features into a single representation. The best performance is achieved when performing multiple segment-to-attribute comparisons.


\begin{table}[!b] 
\centering
\begin{adjustbox}{width=0.8 \textwidth,center}
    \begin{tabular}{ | c ?{0.4mm} c | c ?{0.4mm} c | c | c |}
    \hline
  \multirow{2}{*}{\textbf{Method}}  & \textbf{UCF-101}  & \textbf{UCF-101} & \textbf{HMDB51} \\

&  \textbf{(51/50)}  & \textbf{(81/20)} & \textbf{(26/25)} \\
\Xhline{3\arrayrulewidth}
 \textbf{TARN (w/o attention, single comparison)} &  16.7$\pm$4.0  & 35.8$\pm$5.9 &  16.6$\pm$3.4 \\ \hline
 
 \textbf{TARN (with attention, single comparison)} & 20.0$\pm$2.5  & 38.1$\pm$5.6 &  17.4$\pm$2.8 \\ \hline
 
 \textbf{TARN (with attention, multi comparison)} &  \textbf{23.2}$\pm$\textbf{2.9}  & \textbf{42.7}$\pm$\textbf{5.4} &  \textbf{19.5}$\pm$\textbf{4.2} \\ \hline
 
 

    \end{tabular}
\end{adjustbox}
\caption{Accuracies of the TARN model at different settings on zero-shot action recognition on the UCF-101 and HMDB51 datasets.}
\label{TARN_ZSL}
\end{table}



\textbf{Comparison to state of the art in ZSL:} Methods proposed in the literature for ZSL have used a wide range of testing settings. In order to have a common setting across the majority of works, we do not compare with works that: (1) use auxiliary data to augment the training set \cite{xu2015semantic, xu2016multi, zhu_2018_cvpr}; (2) fuse different semantic or visual features \cite{kodirov2015unsupervised, wang_17_bidilel}; or (3) require access to the testing (unseen) classes during training (also known as ``transductive'' setting) \cite{kodirov2015unsupervised, xu2015semantic, xu2016multi, wang_17_bidilel}. This allows us to have a clear and model-based comparison to the literature. Table \ref{SOA_ZSL} summarizes the comparison results over the UCF-101 (51/50), UCF-101 (81/20), and HMDB51 (26/25) splits. We state in Table \ref{SOA_ZSL} the type of semantic and visual representation used in each method. Our architecture achieves the best results over the UCF-101 51/50 and 81/20 splits with almost 0.5\% and 3\%, respectively, in comparison to the second best performing methods. For HMDB51, we only get lower results than works that use either Improved Dense Trajectories (IDT) as visual features~\cite{zhu_2018_cvpr,qin2017zero}, or ECOC as semantic representation~\cite{qin2017zero}.


\begin{table}[!t] 
\centering
\begin{adjustbox}{width=0.8 \textwidth,center}
    \begin{tabular}{ | c ?{0.4mm} c | c ?{0.4mm} c | c | c |}
    \hline
  \multirow{2}{*}{\textbf{Method}}  & \textbf{Visual} & \textbf{Semantic} & \textbf{UCF-101}  & \textbf{UCF-101} & \textbf{HMDB51} \\
  
           & \textbf{Repr.} & \textbf{Repr.} & \textbf{(51/50)}  & \textbf{(81/20)} & \textbf{(26/25)} \\  
\Xhline{3\arrayrulewidth}
 
 \textbf{ESZSL$\dagger$~\cite{romera2015embarrassingly} - ICML 2015 } & IDT  & WV & 15.0$\pm$1.3  & - & 18.5$\pm$2.0 \\
 \hline
 \multirow{2}{*}{\textbf{SJE$\dagger$~\cite{akata2015evaluation} - CVPR 2015}} & \multirow{2}{*}{IDT} & Attr & 12.0 $\pm$ 1.2  & - & - \\ \cline{3-6}
                                          &     & WV & 9.9 $\pm$ 1.4  & -  & 13.3 $\pm$ 2.4 \\ \hline
 
  \textbf{UDICA \cite{GanYG16} - CVPR 2016} & \multirow{2}{*}{C3D} & \multirow{2}{*}{Attr} & -  &  29.6$\pm$1.2 & - \\ \cline{1-1} \cline{4-6} 
  \textbf{KDICA \cite{GanYG16} - CVPR 2016} &  &  & -  &  31.1$\pm$0.8 & - \\ \hline

 \multirow{2}{*}{\textbf{MTE \cite{xu2016multi} - ECCV 2016}} & \multirow{2}{*}{IDT} & Attr & 18.3$\pm$1.7  & - & - \\  \cline{3-6}
                                           &  & WV & 15.8$\pm$1.3 & - & 19.7$\pm$1.6 \\ \hline
  
 
 \multirow{3}{*}{\textbf{ZSECOC \cite{qin2017zero} - CVPR 2017}} & \multirow{3}{*}{IDT} & Attr & 3.2$\pm$0.7  & - & - \\  \cline{3-6}
                                              &  & WV & 13.7$\pm$0.5  & - & 16.5$\pm$3.9 \\ \cline{3-6}
                                              &  & ECOC & 15.1$\pm$1.7  & - & 22.6$\pm$1.2 \\ \hline
                                              
 \multirow{2}{*}{\textbf{BiDiLEL \cite{wang_17_bidilel} - IJCV 2017}} & \multirow{2}{*}{C3D} & Attr & 20.5$\pm$0.5  & 39.2$\pm$1.0 & - \\  \cline{3-6}
                                               &  & WV & 18.9$\pm$0.4  & 38.3$\pm$1.2 & 18.6$\pm$0.7 \\ \hline
 
 \multirow{2}{*}{\textbf{GMM \cite{mishra_18_generative} - WACV 2018}} & \multirow{2}{*}{C3D} & Attr & 22.7$\pm$1.2  & - & - \\  \cline{3-6}
                                           &  & WV & 17.3$\pm$1.1  & - & 19.3$\pm$2.1 \\ \hline
 
 \textbf{UAR \cite{zhu_2018_cvpr} - CVPR 2018} & IDT & WV &  17.5$\pm$1.6 & - & \textbf{24.4}$\pm$\textbf{1.6} \\ \hline
  \multirow{2}{*}{\textbf{TARN}} & \multirow{2}{*}{C3D} & Attr & \textbf{23.2}$\pm$\textbf{2.9}  & \textbf{42.7}$\pm$\textbf{5.4}  & - \\  \cline{3-6}
                                           &  & WV & 19.0$\pm$2.3  & 36.0$\pm$5.3 & 19.5$\pm$4.2  \\ \hline

    \end{tabular}
\end{adjustbox}
\caption{Accuracies of the TARN model as well as other state-of-the-art methods on zero-shot action recognition on the UCF-101 and HMDB51 datasets. Results marked with ``$\dagger$'' are reported as reproduced by~\cite{xu2016multi}.
}
\label{SOA_ZSL}
\end{table}

\section{Conclusion}

In this work, we propose a deep network (called TARN) for addressing the problem of few-shot and zero-shot action recognition. TARN includes an embedding module for encoding the query and sample set examples, and a relation module that utilize attention for performing temporal alignment and a deep network for learning deep distance measure on the aligned representations at video segment level. The proposed network requires no additional resources or fine-tuning on the target problem. Our experimental results show that using attention and comparing segment-wise representations benefit the video-to-video or video-to-vector matching, compared to using video-wise representations. Furthermore, our method achieves the state-of-the-art results in FSL and very competitive performance in ZSL.



\section{Acknowledgments}

The work of Mina Bishay is supported by the Newton-Mosharafa PhD scholarship, which is jointly funded by the Egyptian Ministry of Higher Education and the British Council. This research has also been supported by EPSRC under grant No. EP/R026424/1. We gratefully acknowledge NVIDIA for the donation of the GTX Titan X GPU used for this research.


\bibliography{bmvc_review_split}
\end{document}